\pdfoutput=1

\documentclass[11pt]{article}

\usepackage[]{emnlp2022}

\usepackage{times}
\usepackage{latexsym}

\usepackage[T1]{fontenc}

\usepackage[utf8]{inputenc}

\usepackage{microtype}

\usepackage{inconsolata}

\usepackage{bbm}
\usepackage{enumitem}
\usepackage{amsmath,bm}
\usepackage{amssymb}
\usepackage{booktabs}
\usepackage{multirow}
\usepackage{graphicx}
\usepackage{xspace}
\graphicspath{ {images/} }

\usepackage[normalem]{ulem}
\useunder{\uline}{\ul}{}

\newcommand{\relu}{\text{ReLU}}

\usepackage[skip=5pt]{caption}
\title{Multilingual Machine Translation with Hyper-Adapters}

\author{Christos Baziotis$^{\star\dagger\bigtriangledown}$ \quad Mikel Artetxe$^{\bigtriangleup}$ \quad James Cross$^{\bigtriangleup}$ \quad Shruti Bhosale$^{\star\bigtriangleup}$ \\
  $^{\bigtriangledown}$ Institute for Language, Cognition and Computation \\
School of Informatics, University of Edinburgh \\
  $^{\bigtriangleup}$ Meta AI Research, Menlo Park, CA, USA \\
}

\begin{document}
\maketitle

\renewcommand* {\thefootnote}{\fnsymbol{footnote}}
\footnotetext{$\star$Correspondence to \href{mailto:c.baziotis@ed.ac.uk}{c.baziotis@ed.ac.uk} or \href{mailto:shru@fb.com}{shru@fb.com}}
\footnotetext{$\dagger$Work done during an internship at Meta AI}
\renewcommand*{\thefootnote}{\arabic{footnote}}

\begin{abstract}

Multilingual machine translation suffers from negative interference across languages. 
A common solution is to relax parameter sharing with language-specific modules like adapters.
However, 
adapters of related languages are unable to transfer information,
and their total number of parameters becomes prohibitively expensive as the number of languages grows.
In this work, we overcome these drawbacks using \textit{hyper-adapters}---hyper-networks that generate adapters 
from 
language and layer embeddings.
While past work had poor results when scaling hyper-networks, 
we propose a rescaling fix that significantly improves convergence 
and enables training larger hyper-networks.
We find that hyper-adapters are more parameter efficient than regular adapters, reaching the same performance with up to 12 times less parameters. When using the same number of parameters and FLOPS, our approach consistently outperforms regular adapters. 
Also, hyper-adapters converge faster than alternative approaches and scale better than regular dense networks.
Our analysis shows that hyper-adapters learn to encode language relatedness, enabling positive transfer across languages.\looseness=-1

\end{abstract}

\section{Introduction}
\label{sec:intro}

Multilingual neural machine translation (MNMT) models~\cite{ha2016multilingual,johnson2017multilingual} 
reduce operational costs and scale to a large number of language pairs~\cite{aharoni-etal-2019-massively}
by using a shared representation space.
This approach benefits low-resource languages through positive transfer from related languages, but introduces a \textit{transfer-interference trade-off}~\cite{arivazhagan2019massively}---as the number of languages grows, 
the performance in more resource-rich languages starts to drop. 
Prior work shows that constrained model capacity 
prevents models from representing all languages equally well~\cite{arivazhagan2019massively}. 
While naively increasing capacity is certain to improve performance~\cite{arivazhagan2019massively,zhang-etal-2020-improving}, it comes with large computational costs.

\begin{figure}[t]
    \centering
    \includegraphics[width=1\columnwidth]{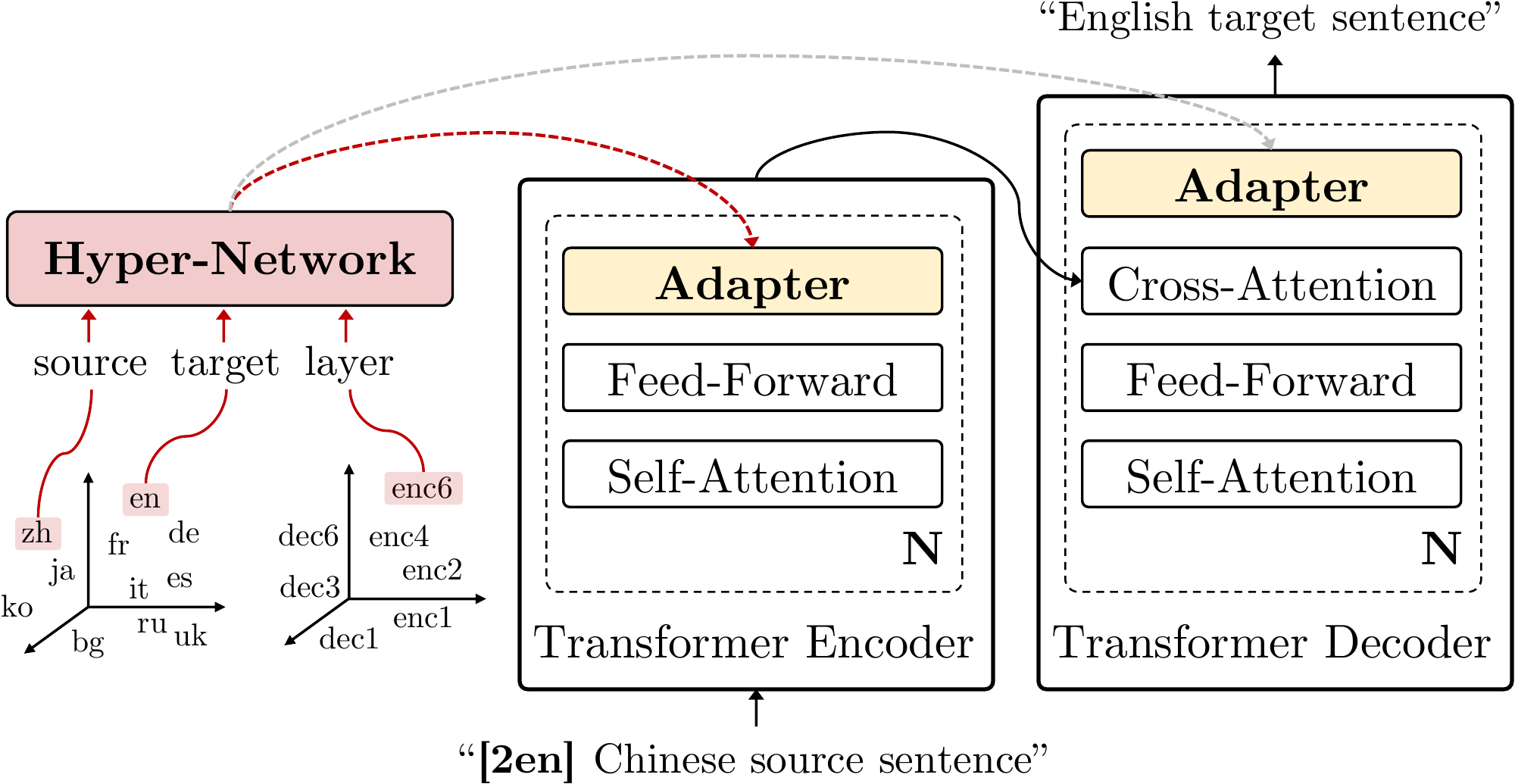}
    \caption{We inject language(-pair)-specific adapters in MNMT, by generating them from a hyper-network.}
    \label{fig:overview}
\end{figure}

A common remedy for the capacity bottleneck is to relax the information sharing with language\allowbreak-specific parameters~\cite{blackwood-etal-2018-multilingual, sachan-neubig-2018-parameter,wang2018multilingual,tan-etal-2019-multilingual,zhang-etal-2020-improving, fan2021beyond}.
Adapter modules~\cite{rebuffi2017adapters}
have been successfully employed in various natural language processing tasks
to address similar capacity-related issues~\cite{houlsby2019parameter, pfeiffer-etal-2020-mad}. 
In MNMT, adapters have been used to adapt (via finetuning) pretrained generic models to specific language-pairs or domains\cite{bapna-firat-2019-simple},
to improve zero-shot performance~\cite{philip-etal-2020-monolingual},
or to reduce interference~\cite {zhu-etal-2021-counter-interference}.

However, using regular language(-pair) adapters has certain limitations. 
First, they can be very parameter-inefficient. 
While each adapter layer might be small, 
the total number of layers is proportional to the number of languages.
This quickly becomes very costly, 
in particular in massively multilingual settings.
In addition, there is no information sharing between the adapters of related languages.
For instance, an adapter for Nepali cannot benefit from the more abundant Hindi data, 
which prevents positive transfer between the two languages.

In this work, we train MNMT models with extra language-specific modules generated by a hyper-network~\cite{ha2017hypernetworks}.
We use adapters for the language-specific modules, dubbed \textit{hyper-adapters}, but it is trivial to replace them with any other architecture.
Hyper-adapters~(Figure~\ref{fig:overview}) are a function of jointly trained language and layer embeddings.
This approach naturally encodes language relatedness and enables knowledge transfer between 
related languages.
It also substantially improves parameter efficiency,
as the number of hyper-adapter parameters is invariant to the number of languages.
We also address optimization obstacles~\cite{sung2021vl}
overlooked by prior work~\cite{karimi-mahabadi-etal-2021-parameter, ansell-etal-2021-mad-g},
and propose a rescaling fix that improves convergence and
enables us to successfully scale to large hyper-networks.

We present experiments on a large multilingual translation benchmark. 
Unlike prior work~\cite{bapna-firat-2019-simple,philip-etal-2020-monolingual} 
that finetunes adapters for language-specific adaptation, 
we train regular- and hyper-adapters jointly with the main network.
We show that with the same parameter budget and FLOPS,  
hyper-adapters are consistently better than other regular adapter variants.
We also match the performance of regular adapters with hyper-adapters up to 12 times smaller.
Hyper-adapters, also converge faster than other approaches
and improve scalability, as small dense networks with hyper-adapters yield similar results to larger regular dense networks.
Our analysis reveals that hyper-adapters do indeed exploit language similarity, 
unlike regular adapters.
By comparing models on benchmarks with artificially constructed properties,
we find that the gains of hyper-adapters grow as the redundancy (e.g., language similarities) in the training data increases.

Our main contributions are:
\setlist[enumerate]{leftmargin=17pt}
\begin{enumerate}
[topsep=3pt,itemsep=3pt,partopsep=0pt, parsep=0pt]
    \item We present a novel approach that injects language-specific parameters in MNMT,
    by generating them from a hyper-network. 
    We also successfully train large hyper-networks by addressing unresolved optimization obstacles.
    \item We present multilingual translation experiments.
    Hyper-adapters consistently outperform regular adapters with the same parameter count or match the results of much larger (up to 12x) regular adapters.
    They also converge faster and scale better than other methods.
    
    \item We present an analysis using a series of probes. 
    We verify that hyper-adapters encode language relatedness, unlike regular adapters. 
    We also find that the gains of hyper-adapters are proportional to the redundancy in the training data.
\end{enumerate}

\section{Background: Multilingual NMT}
\label{sec:mnmt}

In this work, we train universal MNMT models following~\citet{johnson2017multilingual}.
We prepend a special token $\langle2\textsc{XX}\rangle$ to the source and target sequences,
that denotes the target language.
Given a source sentence 
$\bm{x} = \langle x_1, x_2, ..., x_{|\bm{x}|} \rangle$,
a target sequence 
$\bm{y} = \langle y_1, y_2, ..., y_{|\bm{y}|} \rangle$
and a target language token $\bm{t}$, we train our models as follows:
\begin{align*}
    \bm{H} &= \text{encoder}([t, \bm{x}]) \\
    \bm{S} &= \text{decoder}([t, \bm{y}, \bm{H}]) 
\end{align*}

\noindent We use the Transformer architecture~\cite{vaswani2017transformer}
as the backbone of all our models.

\subsection{Language-Specific Parameters}
With universal MNMT, the issue of \textit{negative interference} between unrelated languages emerges, and high-resource language directions are bottlenecked by constrained model capacity ~\cite{arivazhagan2019massively}.
A common solution is to extend model capacity with language-specific modules~\cite{blackwood-etal-2018-multilingual,sachan-neubig-2018-parameter,vazquez-etal-2019-multilingual,wang2018multilingual,lin-etal-2021-learning,zhang-etal-2020-improving,fan2021beyond}.

\paragraph{Adapters}
In this work, we incorporate language-specific parameters using adapter modules,
as they are generic and widely adopted by the 
community for multilingual or multi-task problems.
We follow the formulation of~\citet{bapna-firat-2019-simple, philip-etal-2020-monolingual}, 
and inject one adapter block after each Transformer layer, 
followed by a residual connection.
Let $\bm{z}_i \in \mathbb{R}^{d_z}$ be the output of the $i$-th encoder or decoder layer,
where $d_z$ is the embedding dimension of the Transformer model. 
First, we feed $\bm{z}_i$ to a LayerNorm sublayer $\bar{\bm{z}}_i=\text{\textsc{LN}}_i(\bm{z}_i|\bm{\beta,\gamma})$.
Next, we transform $\bar{\bm{z}}_i$ by applying a down-projection 
$\bm{D_i} \in \mathbb{R}^{d_z \times d_b}$, 
followed by a non-linearity $\bm{\phi}$, 
an up-projection $\bm{U_i} \in \mathbb{R}^{d_b \times d_z}$,
and a residual connection, where $d_b$ is the bottleneck dimensionality of the adapter. 
Formally, each adapter is defined as:
\begin{align*}
    \text{adapter}_i(\bm{z}_i) = \bm{U_i}(\bm{\phi}(\bm{D_i} \, \text{\textsc{LN}}_i(\bm{z}_i))) + \bm{z}_i
\end{align*}
In this work, we use $\relu$ as the non-linearity $\bm{\phi}$.

\paragraph{Adapter Variants}
In MNMT, prior work has used adapters for language(-pair) adaptation, via finetuning.
In our work, we consider two variants 
but train the adapters jointly with the main network. 
Preliminary experiments also showed that jointly training adapters with the main networks yields better results than finetuning adapters.
The first variant is \textit{language-pair adapters}~\cite{bapna-firat-2019-simple},
which uses a different adapter module per language pair in each encoder and decoder layer.
This approach is effective,
but it quickly becomes prohibitively expensive in a multi-parallel setting\footnotemark,
as the number of adapter layers scales quadratically with the number of languages.
Next, we consider (monolingual) \textit{language adapters}~\cite{philip-etal-2020-monolingual},
which use one adapter per language.
Specifically, during xx$\rightarrow$yy translation, 
we activate the adapters for the xx (source) language in the encoder 
and the yy (target) language in the decoder.
Thus, they require fewer adapter layers while also they
generalize to unseen translation directions.

\footnotetext{Multi-parallel refers to a fully many-to-many setting,
unlike the English-centric setting that is \{en$\rightarrow$X $\cup$ X$\rightarrow$en\}.}

\section{Hyper-Adapters}
\label{sec:approach}

\begin{figure}[t]
    \centering
    \includegraphics[width=1\columnwidth]{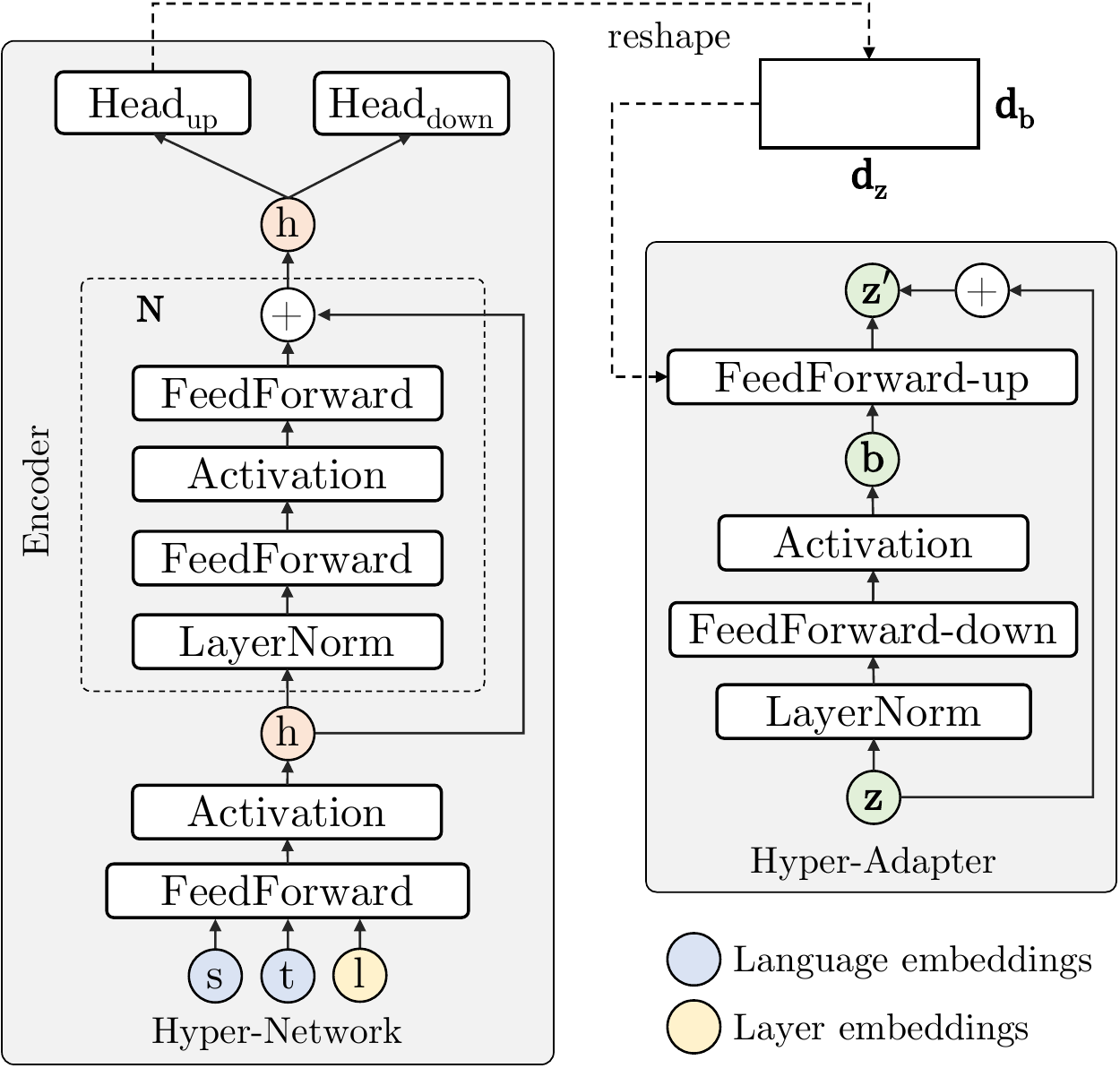}
    \caption{We feed source language, target language 
    and layer-id embeddings into a shared hyper-network,
    to generate adapters weights for all languages and layers.}
    \label{fig:hyper-network}
\end{figure}

We propose to use a hyper-network~\cite{ha2017hypernetworks},
a network that generates the weights of another network,
to produce the weights of all adapter modules, 
dubbed \textit{hyper-adapters}.
As shown in Figure~\ref{fig:hyper-network},
we use a single hyper-network
to generate adapters for all languages and layers
by conditioning on $(\bm{s},\bm{t},\bm{l})$ tuples,
where $\bm{s}$ and $\bm{t}$  denote the source and target language
and $\bm{l}$ denotes the encoder or decoder layer-id (e.g., enc3).
Unlike regular adapters, our approach enables information sharing across languages and layers, and the hyper-network can learn to optimally allocate its capacity across them. Our hyper-network has 3 components:

\paragraph{Input}
We first embed $(\bm{s},\bm{t},\bm{l})$.
We use a shared matrix for the source and target language embeddings,
and a separate matrix for the layer-id embeddings for all encoder and decoder layers.

\paragraph{Encoder}
The language and layer embeddings are given as input to the hyper-network encoder.
First, we concatenate the embeddings and project them with
$\bm{W_{\text{in}}}$, followed by a non-linearity $\bm{\phi}$\footnote{In this work we use $\relu$.}, 
to obtain the hyper-network hidden representation $\bm{h} \in \mathbb{R}^{d_h}$:
\begin{align}
    \bm{h} = \bm{\phi}(\bm{W_{\text{in}}} \, [\bm{s} \| \bm{t} \| \bm{l}])
     \label{eq:1}
\end{align}
\noindent where $\|$ denotes the concatenation operation.
We then pass $\bm{h}$ through $N$ residual blocks,
to encode high-level interactions between the input features:
\begin{align}
    \text{enc}(\bm{h_{i+1}}) = \bm{W_2}(\bm{\phi}(\bm{W_1} \, \text{\textsc{LN}}(\bm{h_i}))) + \bm{h_i}
    \label{eq:2}
\end{align}
where $\bm{W_1}\in\mathbb{R}^{d_h\times d_h}$ and $ \bm{W_2}\in\mathbb{R}^{d_h\times d_h}$
are the trainable weights of each residual block.

\paragraph{Projections}
We feed the final representation $\bm{h}$ to separate projection heads
to obtain (by reshaping their outputs) each weight matrix of a hyper-adapter.
Specifically, 
we use $\bm{H_{\text{up}}} \in \mathbb{R}^{d_h \times (d_b d_z)}$
to generate the weights for each up-projection 
$\bm{U} \in \mathbb{R}^{d_b \times d_z}$,
$\bm{H_{\text{down}}} \in \mathbb{R}^{d_h \times (d_z  d_b)}$
to generate the weights for each down-projection
$\bm{D} \in \mathbb{R}^{d_m \times d_b}$.
We also generate the LayerNorm parameters
$\bm{\gamma} \in \mathbb{R}^{d_z}$ and $\bm{\beta} \in \mathbb{R}^{d_z}$,
with the projection heads 
$\bm{H_{\gamma}} \in \mathbb{R}^{d_h \times d_z}$
and
$\bm{H_{\beta}} \in \mathbb{R}^{d_h \times d_z}$, respectively.

\subsection{Unlocking Large Hyper-networks}
\label{sec:rescaling}

Prior work~\cite{karimi-mahabadi-etal-2021-parameter, ansell-etal-2021-mad-g} in natural language understanding (NLU) 
has used the equivalent of small values of $\bm{d_h}$ and only considered finetuning.
\citet[Fig.~4]{sung2021vl}, recently found that scaling up hyper-networks~\cite{karimi-mahabadi-etal-2021-parameter} 
leads to poor results, 
which they attributed to unknown optimization issues.
In preliminary experiments, we found similar issues when using larger $\bm{d_h}$ values  
(i.e., increasing the hyper-network size). 
We found that the issues were more pronounced when training hyper-adapters jointly with the main network from scratch, 
and speculate that this is a harder optimization problem
than training them with a pretrained and frozen main network.
Next, we identify the cause of this problem and propose a simple fix that allows us to effectively scale hyper-adapters.

\begin{figure}[t]
    \centering
    \includegraphics[width=1\columnwidth]{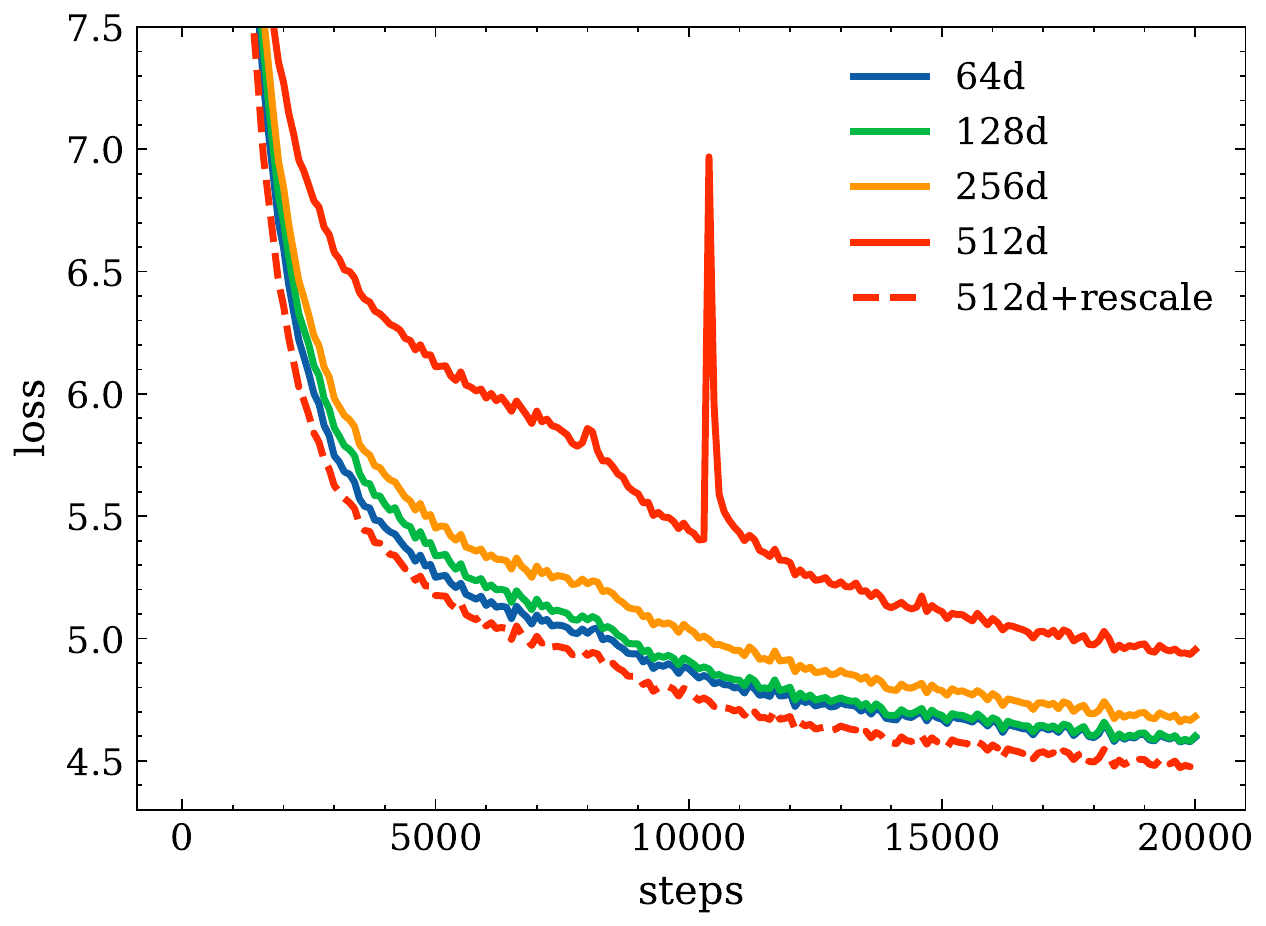}
    \caption{Effect of increasing $\bm{d_h}$ on training. Without rescaling the weights, 
    as we use bigger hyper-networks, training becomes unstable and the loss increases. }
    \label{fig:rescaling-loss}
\end{figure}

\begin{figure}[t]
    \centering
    \includegraphics[width=1\columnwidth]{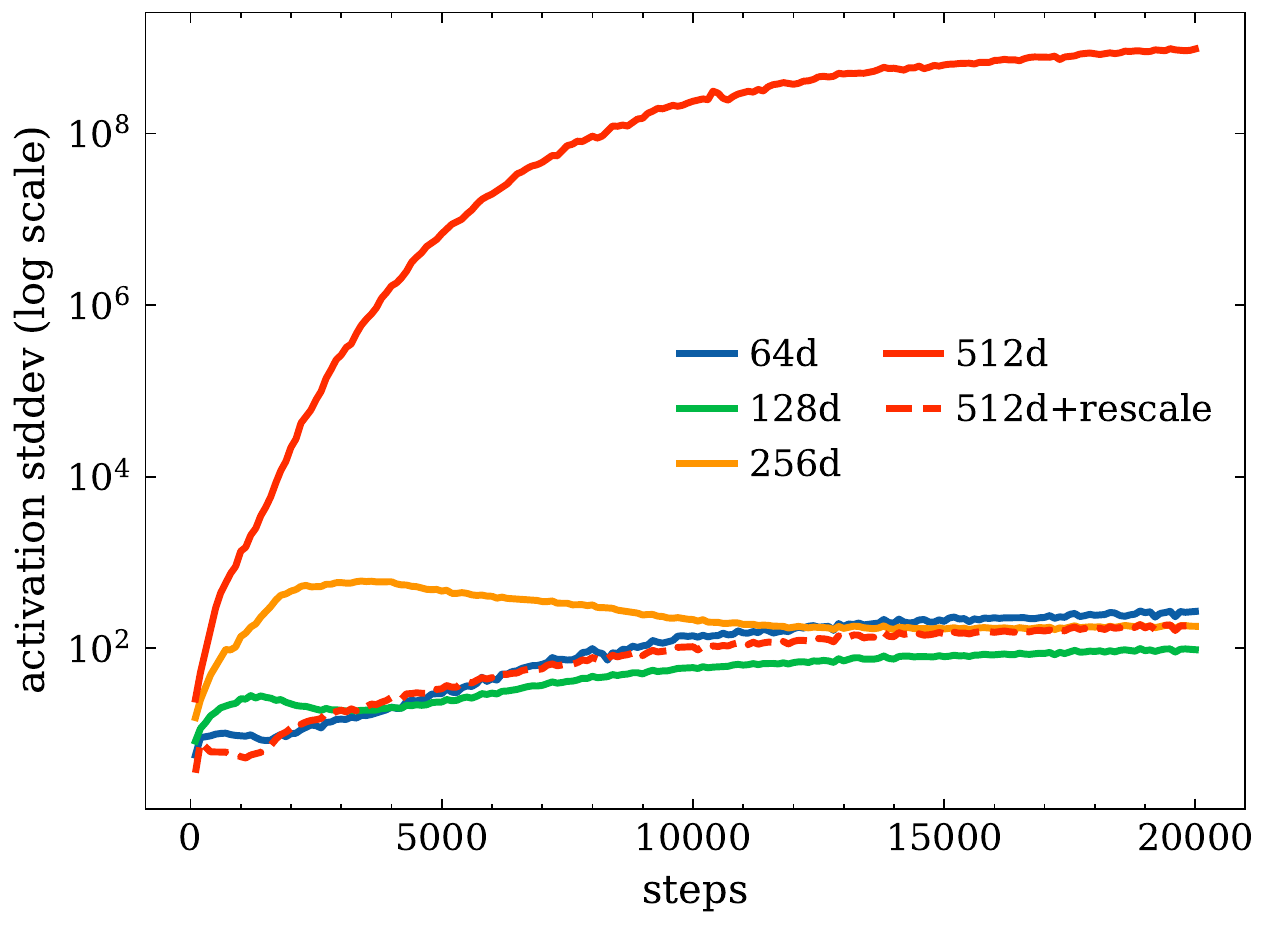}
    \caption{Transformer layer activations as we vary $\bm{d_h}$.}
    \label{fig:rescaling-std}
\end{figure}

Figure~\ref{fig:rescaling-loss} shows the training loss curve as we vary $\bm{d_h}$.
We find that increasing the hyper-network size by increasing $\bm{d_h}$ leads to worse instead of better performance and also makes training very unstable.
In Figure~\ref{fig:rescaling-std}, we plot the average standard deviation (SD) of the Transformer layer activations during training,
and find that for small $\bm{d_h}$, the activations stay within a healthy range,
but as we increase $\bm{d_h}$, the activations start to grow fast.
After a certain point, the network fails to recover
and the activations grow to extreme values.

To solve this issue,
we scale down the generated adapter weights by $\frac{1}{\sqrt{\bm{d_h}}}$,
and generate the adapter weights as 
$\tilde{W} = \text{reshape}(\frac{\bm{H\, \bm{h}}}{\sqrt{\bm{d_h}}})$.
Note that, each component of the generated adapter matrix $\tilde{W}$
is the dot-product of $\bm{h}$ and the corresponding column of 
a given projection head $\bm{H}$.
Thus, the generated weights' SD is proportional to $\bm{d_h}$.
The motivation is similar to the scaled dot-product in Transformer's self-attention.
Once we apply the rescaling fix, 
the activations stay within a healthy range (Figure~\ref{fig:rescaling-std}), 
and increasing $\bm{d_h}$ improves convergence as expected (Figure~\ref{fig:rescaling-loss}).
Note that, in this work we consider variants with $\bm{d_h} > 512$, 
and the rescaling fix is crucial to  unlocking these variants.

\subsection{Parameter Efficiency and FLOPS}
\label{sec:param-efficiency}
Given $\bm{N}$ languages,
language adapters introduce $\bm{N}$ new modules,
whereas language-pair adapters introduce $\bm{N^2}$ new modules in a multi-parallel setting
or $\bm{2N}$ modules in an English-centric many-to-many setting.
By contrast, the number of extra parameters in hyper-adapters is invariant to both
the number of languages and layers.
Most of the parameters are in the projection heads.
Intuitively, 
each row of a head's weight matrix is equivalent to a (flattened) adapter weight matrix.
The number of rows in each head is equal to the hidden size $\bm{d_h}$,
thus $\bm{d_h}$ controls its capacity.
Therefore, to reduce the memory needs compared to language adapters
we must use $\bm{d_h} < \bm{N}$, 
and $\bm{d_h} < \bm{2N}$ for \textit{English-centric} language-pair adapters (details in Appendix~\ref{app:param-efficiency}).

In terms of computational cost, all adapter and hyper-adapter variants yield models with the same FLOPS.
This is because, at test time, we activate only the main network and the corresponding adapters,
with both regular and hyper-adapters having identical architecture and size.
During training, hyper-adapters incur an additional cost for generating the adapter parameters. However, this cost is negligible in practice, as it is run only once per batch for each language pair. At test time, the generated weights can be precomputed and cached.

\begingroup
\setlength{\tabcolsep}{6.0pt} %
\renewcommand{\arraystretch}{1.0} %
\begin{table*}[!tb]
	\small
	\centering

\begin{tabular}{lrrrrrrrrrrr}
\toprule
\multirow{2}[3]{*}{\textbf{Model}} & 
\multicolumn{2}{c}{\textbf{Params}} & 
\multicolumn{4}{c}{\textbf{En→X}} & 
\multicolumn{4}{c}{\textbf{X→En}} & 
\multicolumn{1}{c}{\multirow{2}[3]{*}{\textbf{Mean}}} 
\\ 
\cmidrule(lr){2-3} \cmidrule(lr){4-7} \cmidrule(lr){8-11}
& 
\textbf{Total} & 
\textbf{Extra} & 
\textbf{All} & \textbf{High} & \textbf{Med} & \textbf{Low} & 
\textbf{All} & \textbf{High} & \textbf{Med} & \textbf{Low} &  
 \\
\midrule
Transformer-base & 90M & - & 16.8 & 20.2 & 14.3 & 16.6 & 23.9 & 25.3 & 23.0 & 23.6 & 20.3 \\
\addlinespace
+lang-adapters & 171M & 81M & 18.1 & 21.0 & 15.5 & 18.3 & 25.2 & 26.6 & 24.8 & 24.4 & 21.6 \\
+pair-adapters & 250M & 159M & 18.3 & 21.5 & 16.0 & 18.1 & 24.7 & 25.9 & 24.5 & 23.8 & 21.5 \\
\addlinespace
+hyper-adapters (17\%) & 104M & 14M & 17.9 & 21.5 & 15.4 & 17.5 & 24.9 & 26.3 & 24.1 & 24.5 & 21.4 \\
+hyper-adapters (33\%) & 118M & 27M & 18.5 & 22.0 & 16.0 & 18.2 & 25.3 & 26.6 & 24.5 & \textbf{25.0} & 21.9 \\
+hyper-adapters (100\%) & 173M & 83M & \textbf{19.0} & \textbf{22.2} & \textbf{16.6} & \textbf{18.9} & \textbf{25.7} & \textbf{26.9} & \textbf{25.3} & \textbf{25.0} & \textbf{22.3} \\
\bottomrule

\end{tabular}
    \caption{Results with \textit{Transformer-base} models and (hyper-)adapter bottleneck size of 128.}
	\label{table:random-ml50-base}
\end{table*}
\endgroup

\section{Experimental Setup}
\label{sec:exp-setup}

\paragraph{Data}
We present results on ML50~\cite{Tang2020MultilingualTW},
a multilingual translation dataset with 230M sentences
between English and 50 other typologically diverse languages with data from different domains,
and is larger than comparable publicly available MNMT datasets (e.g., 4x larger than OPUS100;~\citealt{zhang-etal-2020-improving}).
We concatenate the En$\rightarrow$X and X$\rightarrow$En directions,
and group languages based on the amount of their training data into
\textsc{high}~($\geq$1M, 14 languages), \textsc{med}~($\geq$100K, 17 languages) and \textsc{low}~($<$100K, 19 languages).
We use SentencePiece\footnotemark~\cite{kudo-richardson-2018-sentencepiece}
to obtain a joint vocabulary of 90k symbols.
We explored smaller and larger vocabularies but empirically found that 90k strikes a good balance between the number of parameters and translation quality.
Finally, we filter out pairs with more than 250 tokens or with a length ratio over $2.5$.

\footnotetext{We use the \texttt{unigram} model with coverage 0.99995.}
 
\paragraph{Sampling}
To obtain a more balanced data distribution
we use temperature-based sampling~\citep{arivazhagan2019massively}.
Assuming that $p_\textsc{l}$ is the probability that a sentence belongs to language $L$, 
we sample sentences for $L$ with a probability proportional to $p_\textsc{l}^{1/T}$, 
where $T$ is a temperature parameter. 
Larger values of $T$ lead to more even sampling across languages.
During preprocessing, we train SentencePiece with $T$=5.
During training, we set $T$=2 as we observed that with larger values, models overfit on low-resource languages.

\paragraph{Model Configuration} 
We use the Transformer-Base architecture~\cite{vaswani2017transformer} in most of our experiments,
which has 6 encoder and decoder layers, embedding size of 512, 
feed-forward filter size of 2048, 8 attention heads, and 0.1 dropout.
To verify the effectiveness of our approach with more large-scale models,
we also consider an experiment with the Transformer-Big configuration, which uses embedding size of 1024, 
feed-forward filter size of 4096, 16 attention heads, and 0.3 dropout.
In all models, we tie the encoder-decoder embeddings and the decoder output projections~\cite{press-wolf-2017-using,inan2017tying}.
All models are implemented\footnotemark in Fairseq~\cite{ott2019fairseq}.

\footnotetext{\href{https://github.com/cbaziotis/fairseq/tree/hyperadapters/examples/adapters}{github.com/cbaziotis/fairseq -- ``hyperadapters'' branch}}

\paragraph{Optimization}
We use Adam~\cite{kingma2014Adam} with $\beta_1=0.9$, $\beta_2=0.98$, and $\epsilon=10^{-6}$
and regularize models with label smoothing~\cite{szegedy2016rethinking} with $\alpha=0.1$.
We train Transformer-Base models with a learning rate of $0.004$ for 360k steps,
and Transformer-Big models with a learning rate of $0.001$ for 220k,
using a linear warm-up of 8k steps, followed by inverted squared decay. 
All models are optimized with large batches of 256k tokens (8k $\times$ 32 V100 GPUs).
The training time for all models is similar, ranging from 4 to 5 days,
with adapter variants being slightly slower than their dense counterparts.

\paragraph{Evaluation}
During training, we evaluate models every 20k steps and select the checkpoint with the best validation loss, 
aggregated across languages.
At test time, we use beam search of size 5.
We evaluate all models using BLEU~\cite{papineni2002bleu} 
computed with Sacre\textsc{bleu}\footnotemark~\cite{post-2018-call}. 

\footnotetext{\textls[-60]{\fontsize{8.5}{8}\selectfont  BLEU+case.mixed+lang.S-T+numrefs.1+smooth.exp+tok.13a+v1.5.1}}

\paragraph{Baselines}
We compare with strong baselines that incorporate language-specific parameters into MNMT.
We consider two adapter variants that yield significant improvements over dense MNMT models,
namely (monolingual) \textit{language adapters} and \textit{language-pair adapters} and set their bottleneck size to 128. 
Given that ML50 contains 51 languages in total, 
language adapters require 612 adapter modules ($51\times12$), 
whereas language-pair adapters require 1224 (i.e., twice as many).

\paragraph{Hyper-adapter Settings}
We use our proposed hyper-network to generate hyper-adapters 
with \textit{identical} architecture as their regular adapter counterparts.
We consider three hyper-network variants in our experiments: 
\textit{base} ($\bm{d_h}=612$), 
\textit{small} ($\bm{d_h}=204$) 
and \textit{tiny} ($\bm{d_h}=102$).
They contain roughly 100\%, 33\%, and 17\% of the parameters of language adapters\footnotemark, respectively.
We set the size of the language and layer embeddings to 50 and use 2 layers in the hyper-network encoder.

\footnotetext{Or 50\%, 17\% and 8\% w.r.t. language-pair adapters}

\section{Results}
\label{sec:results-base}

\begingroup
\setlength{\tabcolsep}{6.5pt} %
\renewcommand{\arraystretch}{1.0} %
\begin{table*}[!tb]
	\small
	\centering

\begin{tabular}{lrrrrrrrrrrr}
\toprule[1.5pt]
\multirow{2}[3]{*}{\textbf{Model}} & 
\multicolumn{2}{c}{\textbf{Params}} & 
\multicolumn{4}{c}{\textbf{En→X}} & 
\multicolumn{4}{c}{\textbf{X→En}} & 
\multicolumn{1}{c}{\multirow{2}[3]{*}{\textbf{Mean}}} 
\\ 
\cmidrule(lr){2-3} \cmidrule(lr){4-7} \cmidrule(lr){8-11}
& 
\textbf{Total} & 
\textbf{Extra} & 
\textbf{All} & \textbf{High} & \textbf{Med} & \textbf{Low} & 
\textbf{All} & \textbf{High} & \textbf{Med} & \textbf{Low} &  
 \\
\midrule
Transformer-Big & 269M & - & 18.5 & 21.2 & 15.7 & 19.1 & 25.7 & 26.6 & 25.0 & 25.7 & 22.1 \\
\addlinespace
+lang-adapters & 591M & 323M & 19.6 & 22.3 & 17.0 & 20.0 & 27.3 & 28.1 & 27.0 & 27.1 & 23.5 \\
+pair-adapters & 902M & 633M & 20.0 & \textbf{22.9} & 17.5 & 20.3 & 27.0 & 28.2 & 27.0 & 26.3 & 23.5 \\
\addlinespace
+hyper-adapters (tiny) & 323M & 54M & 19.7 & 22.4 & 16.8 & 20.4 & 27.4 & 28.0 & 26.7 & \textbf{27.6} & 23.5 \\
+hyper-adapters (small) & 377M & 108M & 20.0 & 22.8 & 17.2 & 20.5 & \textbf{27.5} & \textbf{28.3} & 26.9 & 27.4 & 23.7 \\
+hyper-adapters (base) & 594M & 325M & \textbf{20.3} & \textbf{22.9} & \textbf{17.6} & \textbf{20.9} & 27.4 & \textbf{28.3} & \textbf{27.1} & 27.2 & \textbf{23.9} \\
\bottomrule[1.5pt]

\end{tabular}
	\caption{BLEU ($\uparrow$) scores of the \textit{Transformer-big} models with (hyper-)adapter bottleneck size of 256.}
	\label{table:random-ml50-big}
\end{table*}
\endgroup

\paragraph{Main Results}
Table~\ref{table:random-ml50-base} shows our main results.
All the reported results are from single runs, 
as MNMT training is computationally expensive. 
However, the results are averaged across 50 languages and 100 translation directions, which makes them robust to noise. 
For completeness, we include the non-aggregate results in the appendix (\S~\ref{app:results}).

Consistent with prior work on fine-tuning adapters for MNMT, 
we find that adding language(-pair) adapters brings substantial improvements across the board~\cite{bapna-firat-2019-simple,philip-etal-2020-monolingual}.
However, the dense (Transformer-Base) baseline has fewer parameters and FLOPS
than all adapter variants.

Hyper-adapters-base 
consistently outperforms both regular adapter variants in all directions,
while having the same parameter count as lang-adapters and half the parameter count of pair-adapters.
We also find that our smaller variants yield very competitive results to regular adapters
while being more parameter efficient.
Hyper-adapters-small 
outperforms both regular adapter variants with fewer parameters,
and hyper-adapters-tiny yields comparable results with only 1/6th and 1/12th 
of the capacity of lang-adapters and pair-adapters, respectively.

In the En$\rightarrow$X directions, 
hyper-adapters-base outperforms lang-adapters by 0.9 BLEU and pair-adapters by 0.7 BLEU.
Interestingly, we see gains even in high-resource settings up to +1.2 BLEU,
although regular adapters have dedicated capacity for these language(-pairs). %
In X$\rightarrow$En, 
hyper-adapter-base has smaller improvements on medium- and high-resource languages,
but we observe improvements of +1.2 BLEU on low-resource languages. 
We hypothesize that the lower improvements on X$\rightarrow$En compared to En$\rightarrow$X are partly due to language specific capacity being more valuable when decoding into many different languages.

\paragraph{Regular Adapters}
We discover interesting trade-offs between the regular adapter variants. 
Pair-adapters are better in En$\rightarrow$X, 
which suggests that it is beneficial to have dedicated capacity 
for encoding the source-side of each En$\rightarrow$X pair.
By contrast, language-adapters are stronger in X$\rightarrow$En.
We believe this is because the (single) decoder-side English adapter benefits from observing all the target-side English data,
unlike the separate X-En adapters 
that see only the target-side English data of each pair.
However, hyper-adapters enjoy the best of both approaches, while being more efficient.

\begin{figure}[t]
    \centering
    \includegraphics[width=1\columnwidth]{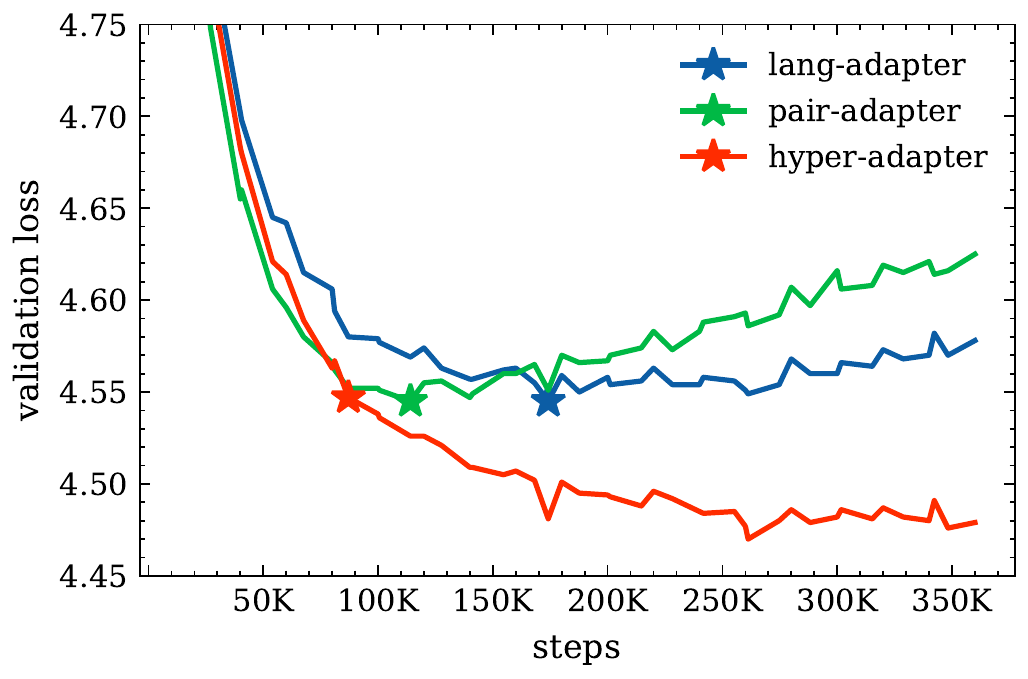}
    \caption{Validation losses of adapter variants.
    We mark when each variant reaches the best loss of lang-adapters.}
    \label{fig:val-loss}
\end{figure}

\paragraph{Convergence}
In Figure~\ref{fig:val-loss},
we compare the validation loss curves of each adapter variant with our hyper-adapters-base variant, which has the same size as lang-adapters.
We mark the point at which each variant reaches the best loss of lang-adapters.
First, we observe that hyper-adapters converge to the best lang-adapters loss at half the number of steps 
(87K-vs-174K).
This shows that assuming a fixed parameter budget, hyper-adapters can significantly reduce training time.
We also find that regular adapters suffer from overfitting, 
in particular pair-adapters.
We suspect this is because using the same capacity for all languages is suboptimal.
\citet{bapna-firat-2019-simple} proposed to use bottleneck sizes proportional to the available training data of a given language pair, which requires tuning.
By contrast, hyper-adapters automatically learn to allocate their available capacity as needed.

\paragraph{Large-Scale Models}
We also evaluate models using the Transformer-Big architecture.
In these experiments, we set the bottleneck size in all adapter and hyper-adapter variants to 256.
We report results in Table~\ref{table:random-ml50-big}.
Overall, we observe similar trends across models as with the Transformer-Base architecture,
although the gains of hyper-adapters are smaller.
We believe this is because we only scale up the main network,
while keeping constant the amount of training data.
Therefore, 
this mitigates the negative interference by reducing the capacity bottleneck, and leaves less room for improvement for language-specific modules, like hyper-adapters.

To our surprise, we find that hyper-adapter-base with the Transformer-Base architecture
(Table~\ref{table:random-ml50-big}) achieves comparable results with the Transformer-Big baseline,
while having significantly fewer parameters (173M-vs-269M)
and a smaller computation cost.
This suggests that hyper-adapters are more effective for addressing negative interference 
than naively scaling up dense networks. 

\section{Analysis}
\label{sec:analysis}
This section investigates why hyper-adapters outperform regular adapters.
Specifically, we focus on how well each adapter variant encodes language relatedness and how it is affected by the redundancy in the training data (i.e., similarities between the data of different languages).
We also explore how modifications in the hyper-network architecture affect
the final model performance.

\begingroup
\setlength{\tabcolsep}{8.0pt} %
\renewcommand{\arraystretch}{1.0} %
\begin{table}[!tb]
\small
\centering
\begin{tabular}{@{}lrrrr@{}}
\toprule[1.5pt]
\textbf{Model}  & 
\textbf{Org} & 
\textbf{Sim} & 
\textbf{Dist} & 
\textbf{Acc $\uparrow$} \\
\midrule
+lang-adapters & 34.1 & 19.0 & 6.5 & 0.56 \\
+pair-adapters & 33.7 & 18.0 & 5.6 & 0.53 \\
+hyper-adapters (base) & 34.8 & \textbf{21.7} & 4.9 & \textbf{0.62} \\
\bottomrule[1.5pt]
\end{tabular}
\caption{BLEU ($\uparrow$) scores of models on the X$\rightarrow$En adapter relatedness probe.
\textit{Org}, \textit{Sim}, \textit{Dist}, 
refer to using the original, similar, and distant source languages, respectively,
while \textit{Acc} denotes the ratio \textit{Sim}/\textit{Org}.}
\label{table:analysis-lang-swap}
\end{table}
\endgroup

\subsection{(Hyper-)Adapter Language Relatedness} 
\label{sec:adapter-relatedness}
We design a probe (Table~\ref{table:analysis-lang-swap}), 
that explicitly compares the ability of regular-vs-hyper adapters to encode language relatedness.
At test time, 
instead of using the adapters of the original source language,
we activate the adapters of another similar, or distant, language.\footnotemark
\footnotetext{For hyper-adapters, we change the source language-id $\bm{s}$.}
We focus on X$\rightarrow$En, 
as we found that changing the target language produced very low BLEU scores, making comparisons unreliable.
We select 4 low-resource languages which have a similar high-resource neighbour in our dataset, 
namely \{af$\rightarrow$nl, pt$\rightarrow$es, gl$\rightarrow$pt, uk$\rightarrow$ru\}.
Also, we consider replacement with ``zh'', 
which is high-resource but distant to all 4 source languages.

When using related languages, 
hyper-adapters suffer less than regular-adapters, 
as they recover more (62\%) of their original BLEU.
Pair-adapters yield worse results than lang-adapters,
presumably due to having weaker target-side (X-En) adapters. 
When using an unrelated language, hyper-adapters suffer the most.
These findings further support that our hyper-networks encode 
language relatedness.

\begingroup
\setlength{\tabcolsep}{4.5pt} %
\renewcommand{\arraystretch}{1.0} %
\begin{table}[!tb]
	\small
	\centering
	\begin{tabular}{@{}lrrrr@{}}
		\toprule[1.5pt]
		\multirow{2}[1]{*}{\textbf{Model}}  & 
		\multicolumn{2}{c}{\textbf{Original}} & 
		\multicolumn{2}{c}{\textbf{Artificial}} \\
        \cmidrule(lr){2-3} \cmidrule(lr){4-5}
		& 
		\multicolumn{1}{c}{\textbf{Param}} & \multicolumn{1}{c}{\textbf{BLEU}} & 
		\multicolumn{1}{c}{\textbf{Param}} & \multicolumn{1}{c}{\textbf{BLEU ($\Delta$)}} \\
		\midrule
        Transformer-Base & 90M & 23.9 & 90M & 23.7 {\color{red} (-0.2)} \\
        +lang-adapters & 114M & 24.7 & 167M & 23.8 {\color{red} (-0.9)} \\
        +pair-adapters & 135M & 24.8 & 240M & 23.9 {\color{red} (-0.9)} \\
        +hyper-adapters & 114M & 24.9 & 114M & 24.9 {\color{blue} (-0.0)} \\
		\bottomrule[1.5pt]
	\end{tabular}
	\caption{BLEU ($\uparrow$) scores on the original-vs-artificial ML15 splits.}
	\label{table:analysis-artificial}
\end{table}
\endgroup

\subsection{The Role of Data Redundancy}
\label{sec:data-redundancy}

We have hypothesized that our hyper-network exploits similarities (i.e., redundancies) in the data,
to produce similar adapters for similar languages and avoid encoding redundant features.
This implies that hyper-adapters would ``degenerate'' into regular adapters 
if the training data contained only distant languages.
To test this hypothesis, we create two different splits out of ML50, 
with and without similar languages.
First, we select 14 (+English) relatively unrelated languages and create ML15\footnotemark. 
Then, we create another version of ML15, 
that emulates a dataset with similar languages.
We split the data of each language into smaller parts (e.g., $\text{fr}_1, \text{fr}_2, \ldots, \text{fr}_N$) 
which we treat as different languages, 
which results in 47 artificial languages.

Table~\ref{table:analysis-artificial} shows the results.
We observe that in the original ML15 version,
regular- and hyper-adapters achieve similar results.
In contrast, in the fragmented ML15 version,
regular adapters suffer significantly as they cannot share information, 
unlike hyper-adapters that are unaffected.
These findings show that the gains of hyper-adapters are 
proportional to the redundancies in the training data.
Thus, we expect that the gap between regular- and hyper-adapter 
will grow as the number of related languages (or their data) grows.
Note that, as the artificial ML15 has more languages, 
regular adapters require more layers and thus more parameters.

\footnotetext{The languages of ML15 are \{en, fr, zh, hi, lt, iu, et, ro, nl, it, ar, tr, km, vi, uk\}.
We include more details in Appendix~\ref{sec:ml15-stats}.}

\begin{figure}[t]
    \centering
    \includegraphics[width=1\columnwidth]{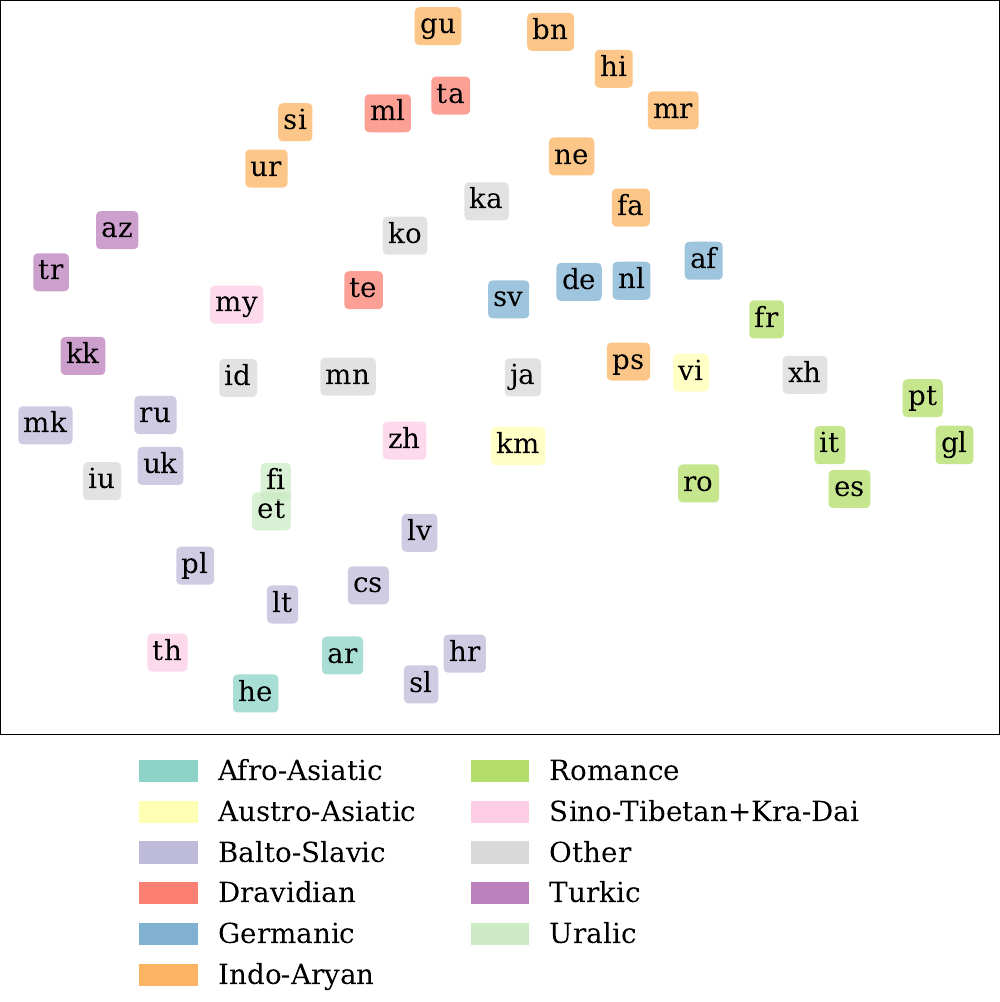}
    \caption{Plot of hyper-network language embeddings.}
    \label{fig:lang-embeddings-umap}
\end{figure}

\subsection{Hyper-Network Embeddings} 
\label{sec:emb-umap}
In Figure~\ref{fig:lang-embeddings-umap}, we visualize the language embeddings 
of our hyper-adapters-tiny variant using UMAP~\cite{mcinnes2018umap}.
We observe that the hyper-network embeds languages that belong to the same family close to each other.
This is another piece of evidence that hyper-adapters encode language relatedness.

\begingroup
\setlength{\tabcolsep}{5.5pt} %
\renewcommand{\arraystretch}{1.1} %
\begin{table}[!tb]
	\small
	\centering
	\begin{tabular}{lrrr}
		\toprule[1.5pt]
		\textbf{Model}                     & 
		\multicolumn{1}{c}{\textbf{En→X}} & 
		\multicolumn{1}{c}{\textbf{X→En}} & 
		\multicolumn{1}{c}{\textbf{Mean}} \\
		\midrule
		Linear                             & 17.3 & 24.4 & 20.8 \\
		Non-Linear                         & 18.2 & 25.0 & 21.6 \\
		Non-Linear + 2 ResBlocks            & 18.5 & 25.2 & 21.8 \\
		\bottomrule[1.5pt]
	\end{tabular}
	\caption{Comparison of encoding methods with Transformer-base models trained for 160K steps.}
	\label{table:analysis-depth}
\end{table}
\endgroup
\subsection{Hyper-Network Encoder} 
In Table~\ref{table:analysis-depth}, 
we compare different methods for encoding the hyper-network inputs $(\bm{s},\bm{t},\bm{l})$
for obtaining the hyper-network output representations $\bm{h}$ (i.e., before generating the hyper-adapter weights).
We find that using only one linear layer is suboptimal, 
and stacking multiple non-linear layers is important.
Specifically, adding a non-linearity to Eq.~\ref{eq:1} (we used ReLU) 
improves performance, 
and stacking more layers helps even further (Eq.~\ref{eq:2}).
We speculate this allows the input features to better interact with each other.
In preliminary experiments, we found that stacking more than 2 layers did not produce consistent improvements.

\begingroup
\setlength{\tabcolsep}{3.5pt} %
\renewcommand{\arraystretch}{1} %

\newcommand{\srctgt}{$(\bm{s},\bm{t})$ }
\newcommand{\srcxxx}{$(\bm{s})$\phantom{+x}}
\newcommand{\tgtxxx}{$(\bm{t})$\phantom{+x}}

\begin{table}[!tb]
	\small
	\centering
    \begin{tabular}{lrrrr}
	\toprule[1.5pt]
	\multirow{2}{*}{\textbf{Model}} &
	\multicolumn{2}{c}{\textbf{Supervised}} &
	\multicolumn{2}{c}{\textbf{Zero-Shot}} \\
	\cmidrule(lr){2-3}\cmidrule(lr){4-5}
	&
	\multicolumn{1}{c}{\textbf{En$\rightarrow$X}} &
	\multicolumn{1}{c}{\textbf{X$\rightarrow$En}} &
	\multicolumn{1}{c}{\textbf{Direct}} &
	\multicolumn{1}{c}{\textbf{Pivot}} \\
	\midrule
	Transformer-Base                            & 16.2          & 23.1          & 12.3          & 16.5          \\
	+lang-adapters                   & 17.9          & 24.9          & 13.1          & 18.8          \\
	\midrule
	+ \textit{hyper-adapters (base)}        &               &               &               &               \\
	\addlinespace
	enc=\srctgt dec=\srctgt          & 18.5          & 25.1          & 1.7           & 19.4          \\
	\addlinespace
	enc=\srctgt dec=\tgtxxx              & \textbf{18.6} & \textbf{25.2} & 8.7           & \textbf{19.6} \\
	\hspace{0.5cm} + dropout=0.1     & 18.3          & 25.0          & 11.4          & 19.3          \\
	\hspace{0.5cm} + dropout=0.2     & 18.1          & 25.0          & 11.7          & 18.8          \\
	\addlinespace
	enc=\srcxxx dec=\tgtxxx     & 17.8          & \textbf{25.2} & \textbf{13.8} & 19.1          \\
	\hspace{0.5cm} + dropout=0.1     & 17.7          & 25.1          & 13.7          & 18.7          \\
	\hspace{0.5cm} + dropout=0.2     & 17.5          & 24.8          & 12.9          & 18.8          \\
	\bottomrule[1.5pt]
    \end{tabular}
	\caption{Effect of different hyper-network input combinations on zero-shot translation. 
	The layer embedding $\bm{l}$ is always used and is omitted for brevity.}
	\label{table:analysis-zero}
\end{table}
\endgroup

\subsection{Zero-Shot Translation} 
\label{sec:zero-shot}
In this analysis (Figure~\ref{fig:lang-embeddings-umap}), 
we investigate the zero-shot capabilities of different hyper-adapter variants.
Specifically, 
we mask either the source or target language in the hyper-network's input $(\bm{s},\bm{t},\bm{l})$
when generating the encoder or decoder hyper-adapters.
We train models for 160k steps to reduce training time.
This means that hyper-adapter haven't fully converged (Figure~\ref{fig:val-loss}), unlike regular adapters.
However, we are interested in comparing different hyper-adapter variants to each other and include lang-adapters only for context. 

\paragraph{Test Data}
To compute the zero-shot results we use the 15 translation combinations between Arabic, Chinese, Dutch, French, German, and Russian, following the setup of~\citet{zhang-etal-2020-improving}.
We use the devtest splits from the FLORES-200 multi-parallel evaluation benchmark~\cite{goyal-etal-2022-flores, nllb2022}.
Each test set contains 3001 sentences from Wikipedia articles.
We evaluate models both in direct zero-shot (i.e., X$\rightarrow$Y) and pivot zero-shot through English (i.e., X$\rightarrow$En$\rightarrow$Y).
Note that pair-adapters cannot do direct zero-shot translation by definition.

\paragraph{Results}
Hyper-adapters fail at direct zero-shot translation 
when using both $\bm{s}$ and $\bm{t}$ in the hyper-network for both the encoder and decoder hyper-adapters.
Masking $\bm{s}$ in decoder hyper-adapters yields a significant boost,
which is further increased by masking $\bm{t}$ in encoder hyper-adapters.
This reveals a trade-off between supervised and zero-shot translation.
Removing the target language information from encoder hyper-adapters harms En$\rightarrow$X translation, 
which is reflected in the pivot-based zero-shot translation.
However, removing the source language information from decoder hyper-adapters 
has no effect on supervised translation, although it improves zero-shot.
These results suggest that the ``enc=$(\bm{s},\bm{t})$ dec=$(\bm{s},\bm{t})$'' variant behaves similar to language-pair adapters, which cannot do zero-shot, 
whereas the ``enc=$(\bm{s})$ dec=$(\bm{t})$'' variant  behaves similar to language-adapters.
In our experiments, we use the ``enc=$(\bm{s},\bm{t})$ dec=$(\bm{t})$'' variant,
which strikes a good balance.

We also explore adding dropout inside the hyper-network layers, 
to produce more robust representations $\bm{h}$, 
but not in the generated hyper-adapters.
We observe small negative effects in the supervised setting, 
but mixed results in the zero-shot setting.
In particular, in the ``enc=$(\bm{s},\bm{t})$ dec=$(\bm{t})$'' variant, 
dropout significantly improves zero-shot.
These results suggest that there is room for improvement in this direction, 
but we leave this for future work.

\section{Related Work}
\label{sec:related}

\citet{platanios-etal-2018-contextual} 
explored an idea similar to hyper-networks in MNMT
with the so-called ``contextual parameter generation'' to promote information sharing across languages,
by generating the weights of an RNN-based~\cite{Bahdanau2014} MNMT model from language embeddings.
By contrast, we consider a hybrid approach that generates only a few (language-specific) modules,
instead of generating all the layers of a Transformer model,
which introduces a large computational overhead.

Another approach is combining hyper-networks with pretrained models.
In NLU,
\citet{karimi-mahabadi-etal-2021-parameter} generate task-specific adapters from task embeddings.
\citet{tay2021hypergrid} use a hyper-network 
to learn grid-wise projections for different tasks.
\citet{ye-ren-2021-learning} 
extend text-to-text Transformers~\cite{raffel2020t5} to unseen tasks 
by generating adapters from task descriptions.

In multilingual dependency parsing, 
\citet{ustun-etal-2020-udapter} generate adapters for the biaffine attention 
from language representations in linguistic databases.
\citet{ansell-etal-2021-mad-g} also use linguistic databases for cross-lingual NLU,
and extend~\citet{pfeiffer-etal-2020-mad} by generating language adapters for unseen languages.
In concurrent work, \cite{ustun2022hyper} consider a conceptually similar approach to our work for multi-task multilingual transfer in NLU tasks.

Unlike prior work,
(1) we identify and solve optimization issues overlooked by other hyper-network-based methods,
(2) we train regular- and hyper-adapters jointly with the main network instead of using them for finetuning,
and (3) we focus on NMT, which is a more complex generation task instead of the relatively simpler NLU tasks.

\section{Conclusion}

In this work,
we extend the capacity of MNMT models with hyper-adapters,
which are language-specific adapter modules generated from a hyper-network.
By resolving optimization issues not addressed by prior work,
we successfully train large hyper-networks from scratch jointly with the rest of the main network on MNMT (\S\ref{sec:rescaling}).

We show that hyper-adapters consistently outperform other regular adapter variants 
across translation directions and model sizes (\S\ref{sec:results-base}),
while improving parameter efficiency. 
We also observe computational efficiency gains, 
as a smaller Transformer-Base model with hyper-adapters gives similar results to the dense Transformer-Big model, which is computationally more expensive and requires more parameters. 
Besides improvements in translation quality, 
hyper-adapters achieve faster training convergence as shown in \S\ref{sec:results-base}.
Our analysis shows that, unlike regular adapters, 
hyper-networks enable positive transfer across the hyper-adapters of similar languages,
by encoding language relatedness (\S\ref{sec:adapter-relatedness},\ref{sec:emb-umap}) 
and exploiting redundancies (i.e., language similarities) in the training data (\S\ref{sec:data-redundancy}).
Finally, by manipulating the input of the hyper-network
we discover that there is a trade-off between the zero-shot and supervised translation performance of hyper-adapters (\S\ref{sec:zero-shot}).

\section*{Limitations}
\paragraph{Modeling} 
As mentioned in Section~\ref{sec:param-efficiency}, 
one limitation of hyper-adapters, compared to regular adapters, 
is that they introduce a small computational overhead during training.
Specifically, in each batch, 
we need to do one pass through the hyper-network to generate the hyper-adapter parameters.
This cost is proportional to the size of the hyper-network and the total number of transformer layers (and thus the hyper-adapter layers to generate).
In this work, we found that this cost was negligible, 
as the number of total transformer layers is small (12) 
and 2-layer deep hyper-network was sufficient to obtain good results.
Besides, the parameter generation cost only affects training time.
Once the training is completed we can pre-generate 
and cache all the hyper-adapter modules,
thus obtaining identical inference cost with regular adapters.

\paragraph{Data}
In our experiments, we use only the ML50 dataset, which is relatively larger than those used in prior works.
However, ML50 contains only English-centric parallel data and real-world multilingual datasets can be much larger, noisy, and diverse~\cite{nllb2022, bapna2022building}.

\section*{Acknowledgments}
We thank Angela Fan, Myle Ott, Vedanuj Goswami, and Naman Goyal for all their help and advice during this project.

\bibliographystyle{acl_natbib}
\bibliography{anthology,custom}

\clearpage
\appendix

\section{ML15 Dataset}
\label{sec:ml15-stats}
\begingroup
\setlength{\tabcolsep}{13pt} %
\renewcommand{\arraystretch}{1.1} %
\begin{table}[!tb]
\small
\centering
\begin{tabular}{lrc}
\toprule[1.5pt]
\textbf{Language} & \multicolumn{1}{l}{\textbf{Sentences}} & \textbf{\# Splits} \\
\midrule
fr\_XX & 38,507,539 & 5 \\
zh\_CN & 11,173,646 & 5 \\
hi\_IN & 1,450,114 & 5 \\
lt\_LT & 1,402,892 & 5 \\
iu\_CA & 1,109,076 & 5 \\
et\_EE & 1,064,974 & 5 \\
ro\_RO & 600,019 & 5 \\
nl\_XX & 232,038 & 2 \\
it\_IT & 226,385 & 2 \\
ar\_AR & 225,678 & 2 \\
tr\_TR & 203,702 & 2 \\
km\_KH & 183,934 & 2 \\
vi\_VN & 127,117 & 1 \\
uk\_UA & 104,021 & 1 \\
\bottomrule[1.5pt]
\end{tabular}
\caption{Statistics of the Ml15 dataset, include the number of splits per language in the artificial version.}
\label{table:ml15-stats}
\end{table}
\endgroup

In Table~\ref{table:ml15-stats}, we show the statistics of the ML15 dataset.
It includes 14 (+English) medium- to high-resource languages from ML50, 
that are relatively distant from each other.
We also create another version of this dataset,
in which we split each language into smaller parts and consider each one of them as a different language.
We aim to have approximately 100k sentences per split with at most 5 splits per language, 
to prevent the number of artificial languages from becoming too large.

\begin{figure}[t]
    \centering
    \includegraphics[width=1\columnwidth]{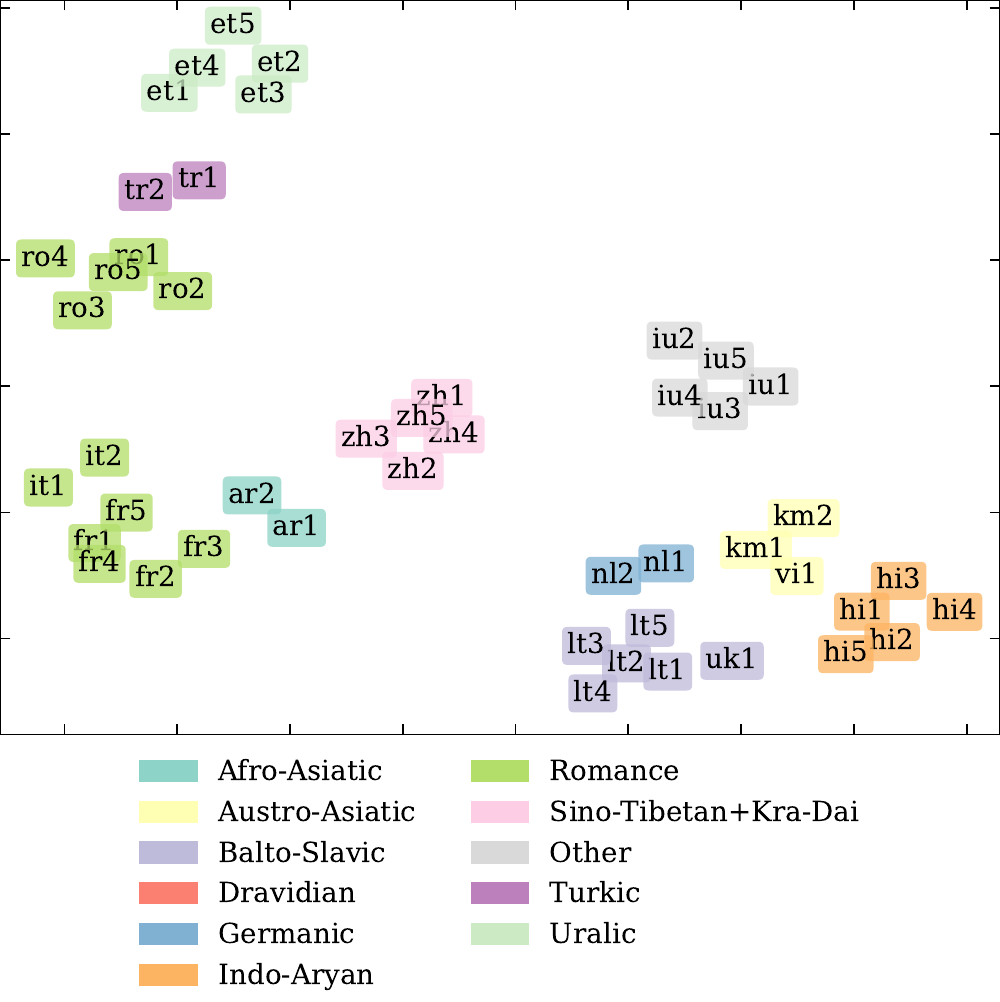}
    \caption{Plot of hyper-network language embeddings trained on the fragmented version of ML15 with the artificial languages.}
    \label{fig:lang-embeddings-ml15-umap}
\end{figure}

In Figure~\ref{fig:lang-embeddings-ml15-umap}, 
we visualize the hyper-network language embeddings 
of the model trained on the fragmented version of ML15 with the artificial languages.
The plot clearly demonstrates that the hyper-network is able to capture the fact that all the artificial splits of a given language are similar to each other.
Based on that, 
it is able to avoid relearning the same features, while also exploiting the all the available (related) data
to learn to generate more powerful hyper-adapters.

\section{Hyper-Network Architecture}

\subsection{Initialization}
\label{sec:hyper-init}

Classical weight initialization methods~\cite{glorot10init,he2015delving}, 
when used to initialize a hyper-network, 
fail to produce weights for the main network on the correct scale.
We explore a simple and generic solution to properly initialize each hyper-network head.

First, we initialize a given projection head $\bm{H}$ with 
a regular initialization method.
Then, we also randomly initialize another (temporary) weight matrix, 
with the same dimensions as the adapter matrix we want to generate from $\bm{H}$, 
and compute its standard deviation $\bm{\sigma_a}$.
Recall that, we want $\bm{H}$ to generate adapter weights in the target scale of $\bm{\sigma_a}$.
Next, we feed a random input $(\bm{s},\bm{t},\bm{l})$ into the hyper-network, 
generate the hyper-adapter weight matrix, 
and compute its standard deviation $\bm{\sigma_h}$. 
Finally, we re-scale the original weights of $\bm{H}$ as
\begin{align*}
    \bm{H'} = 
    \bm{H} \odot \frac{\bm{\sigma_a}}{\bm{\sigma_h}}
\end{align*}
which ensures that the next time the projection head $\bm{H}$ will generate a weight matrix,
it will have values within the desired scale.
In our experiments, we found that this hyper-network aware initialization was helpful 
only when not using our proposed re-scaling (\S~\ref{sec:rescaling}).
However, once we employ the re-scaling, 
all models converge to the same results regardless of initialization.

\citet{Chang2020Principled} first pointed out the importance of properly initialing a hyper-network. 
However, while their proposed initialization is principled, 
it requires to be computed analytically for each source$\rightarrow$target layer mapping. 
By contrast, our method simply initializes a target layer and numerically adjusts the weights of the hyper-network, 
which works for arbitrary layer architectures.

\begingroup
\setlength{\tabcolsep}{6.5pt} %
\renewcommand{\arraystretch}{1.0} %
\begin{table*}[!tb]
	\small
	\centering

\begin{tabular}{lrrrrrrrrrrr}
\toprule[1.5pt]
\multirow{2}[3]{*}{\textbf{Model}} & 
\multicolumn{2}{c}{\textbf{Params}} & 
\multicolumn{4}{c}{\textbf{En→X}} & 
\multicolumn{4}{c}{\textbf{X→En}} & 
\multicolumn{1}{c}{\multirow{2}[3]{*}{\textbf{Mean}}} 
\\ 
\cmidrule(lr){2-3} \cmidrule(lr){4-7} \cmidrule(lr){8-11}
& 
\textbf{Total} & 
\textbf{Extra} & 
\textbf{All} & \textbf{High} & \textbf{Med} & \textbf{Low} & 
\textbf{All} & \textbf{High} & \textbf{Med} & \textbf{Low} &  
 \\
\midrule
Transformer-Base & 90M   & - & 41.7 & 45.3 & 40.6 & 40.1 & 47.8 & 53.0 & 46.8 & 44.7 & 44.7\\
\addlinespace
+lang-adapters & 171M    & 81M & 43.1 & 46.2 & 42.0 & 41.8 & 49.2 & \textbf{54.4} & 48.7 & 45.8 & 46.1\\
+pair-adapters & 250M    & 159M & 43.6 & 46.8 & 42.9 & 42.0 & 48.9 & 54.2 & 48.7 & 45.3 & 46.3\\
\addlinespace
+hyper-adapters (tiny)   & 104M & 14M & 42.7 & 46.5 & 41.5 & 40.9 & 48.6 & 53.9 & 47.7 & 45.6 & 45.7\\
+hyper-adapters (small)  & 118M & 27M & 43.5 & 46.9 & 42.4 & 41.9 & 48.9 & 54.2 & 48.0 & 45.8 & 46.2\\
+hyper-adapters (base)   & 173M & 83M & \textbf{44.2} & \textbf{47.2} & \textbf{43.3} & \textbf{42.8} & \textbf{49.4} & 54.3 & \textbf{48.8} & \textbf{46.3} & \textbf{46.8}\\
\bottomrule[1.5pt]

\end{tabular}
    \caption{ChrF ($\uparrow$) scores of the \textit{Transformer-base} models with (hyper-)adapter bottleneck size of 128.}
	\label{table:random-ml50-base-chrf}
\end{table*}
\endgroup

\subsection{LayerNorm Generation}
\label{sec:hyper-layernorm}
In regular LayerNorm, 
we initialize the scaling parameters $\gamma$ with 1 and the shifting parameters $\beta$ with 0. 
However, when we generate $\tilde{\gamma}$, $\tilde{\beta}$ from the hyper-network, 
their values (initially) will be zero-mean as they are the activations of a randomly initialized projection.
This can cause convergence issues, 
because if the values of $\tilde{\gamma}$ are close to zero, 
then the inputs $z$ would be scaled down close to zero, 
thus slowing down convergence. 
To address this issue, 
we increment the generated weights for $\tilde{\gamma}$ by +1, 
to ensure that they have the desired scale:
\begin{align*}
    \text{\textsc{LN}}(\bm{z}_i | \bm{\tilde{\gamma}}, \bm{\tilde{\beta}}) = \frac{\bm{z}_i - \bm{\mu_{z_i}}}{\bm{\sigma_{z_i}}}  \odot (\bm{\tilde{\gamma}}{\color{red}\bm{+\mathbbm{1}}}) + \bm{\tilde{\beta}}
\end{align*}
where $\bm{\mathbbm{1}}$ denotes a vector of ones.

\subsection{Parameter Efficiency}
\label{app:param-efficiency}

In this section, we discuss the parameter efficiency of each (hyper-)adapter variant in greater detail.
For brevity, we ignore the (negligible) LayerNorm parameters. 

\paragraph{Regular Adapters}
Each adapter block has an up- and a down-projection with equal parameters and total capacity $ C_{\text{block}} = d_z  d_b + d_b d_z = 2 d_z d_b$. 
Language adapters add $C_{\text{lang}} = N \cdot L \cdot C_{\text{block}}$ new parameters into an MNMT model, 
where $\bm{N}$ is the number of languages and $L$ the number of  (encoder+decoder) layers.
Language-pair adapters, introduce  $C_{\text{pair}} = N^2 \cdot L \cdot C_{\text{block}}$ new parameters in a multi-parallel many-to-many setting,
or $C_{\text{pair}} = 2N \cdot L \cdot C_{\text{block}} = 2 \cdot C_{\text{lang}}$ 
new parameters in an English-centric\footnotemark~many-to-many setting.

\footnotetext{Concatenation of English$\rightarrow$X and X$\rightarrow$English directions.}

\paragraph{Hyper-Adapters}
A benefit of hyper-adapters, 
is that their number of parameters is invariant to both
the number of languages $N$ and layers $L$.
Most of the parameters are in the projection heads of the hyper-network.
Intuitively, 
each row of a head's weight matrix is equivalent to a (flattened) adapter weight matrix.
The number of rows in a head is equal to hidden size $\bm{d_h}$ of the hyper-network.
Therefore, $\bm{d_h}$ controls the hyper-network capacity:
\begin{align*}
    C_{\text{hyper}} = 
    \underbrace{\bm{d_h} (d_z d_b)}_{\text{Head-down}} +  
    \underbrace{\bm{d_h} (d_b d_z)}_{\text{Head-up}}
    =  \bm{d_h} \cdot C_{\text{block}}  
\end{align*}
For example, 
in a dataset with $N=50$ languages with a Transformer model with total $L=12$ layers, language adapters introduce 600 adapter blocks.
If we set $\bm{d_h}=600$, then hyper-adapters introduce the same number of parameters,
whereas using $\bm{d_h} < N \cdot L$ yields parameter savings.
The parameter savings with respect to language adapters are $\frac{d_h}{N \cdot L}$, 
and to \textit{English-centric} pair-adapters are $\frac{d_h}{2N \cdot L}$.
The hyper-network embeddings and encoder contain a comparatively negligible amount of parameters.

\section{Additional Results}
\label{app:results}

This section contains additional results for the main experiments in Section~\ref{sec:results-base} with the Transformer-Base models.
Table~\ref{table:random-ml50-base-chrf} shows results measured with ChrF~\cite{popovic-2015-chrf}.
Overall, we observe that the results are consistent with the BLEU scores reported in 
Table~\ref{table:random-ml50-base} in the main paper.
In Tables~\ref{table:random-ml50-base-en-many} and~\ref{table:random-ml50-base-many-en},
we report the non-aggregated BLEU scores for the en$\rightarrow$X and X$\rightarrow$en pairs, respectively.

\begingroup
\setlength{\tabcolsep}{8.5pt} %
\renewcommand{\arraystretch}{1.2} %
\begin{table*}[!tb]
	\small
	\centering
\begin{tabular}{lrrrrrr}
\toprule[1.5pt]

\multicolumn{1}{c}{\multirow{2}[2]{*}{\textbf{Language}}} & 
\multicolumn{1}{c}{\multirow{2}[2]{*}{\textbf{Transformer-Base}}} & 
\multicolumn{5}{c}{\textbf{+adapters}} \\ 
\cmidrule(l){3-7} 
\multicolumn{1}{c}{} & 
\multicolumn{1}{c}{} & 
\textbf{lang} & 
\textbf{pair} & 
\textbf{hyper (base)} & 
\textbf{hyper (base)} & 
\textbf{hyper (base)} \\ \midrule

en$\rightarrow$af & 17.1 & 16.1 & 15.7 & 16.2 & 15.6 & 15.4 \\
en$\rightarrow$ar & 11.5 & 13.4 & 14.0 & 12.5 & 13.3 & 14.0 \\
en$\rightarrow$az & 6.8 & 7.1 & 7.3 & 7.9 & 8.1 & 7.5 \\
en$\rightarrow$bn & 11.2 & 13.0 & 12.5 & 11.0 & 11.9 & 12.5 \\
en$\rightarrow$cs & 19.6 & 20.6 & 20.7 & 20.8 & 20.8 & 21.2 \\
en$\rightarrow$de & 34.3 & 35.2 & 36.0 & 36.1 & 36.6 & 36.8 \\
en$\rightarrow$es & 26.6 & 28.9 & 28.5 & 28.6 & 29.1 & 29.3 \\
en$\rightarrow$et & 15.8 & 16.8 & 17.5 & 17.0 & 17.1 & 17.8 \\
en$\rightarrow$fa & 14.3 & 15.2 & 15.8 & 14.9 & 15.2 & 16.2 \\
en$\rightarrow$fi & 16.8 & 17.7 & 18.7 & 17.8 & 18.9 & 18.9 \\
en$\rightarrow$fr & 34.1 & 34.4 & 35.1 & 35.0 & 35.4 & 35.2 \\
en$\rightarrow$gl & 25.0 & 26.0 & 23.3 & 25.8 & 26.8 & 26.3 \\
en$\rightarrow$gu & 0.4 & 0.4 & 0.3 & 0.2 & 0.1 & 0.2 \\
en$\rightarrow$he & 22.6 & 25.2 & 26.5 & 24.1 & 25.4 & 27.2 \\
en$\rightarrow$hi & 14.9 & 15.8 & 16.9 & 16.5 & 16.7 & 17.1 \\
en$\rightarrow$hr & 25.0 & 28.1 & 28.5 & 27.0 & 28.4 & 29.2 \\
en$\rightarrow$id & 29.6 & 32.4 & 32.3 & 31.5 & 32.5 & 33.2 \\
en$\rightarrow$it & 30.1 & 32.0 & 32.7 & 31.8 & 32.9 & 34.3 \\
en$\rightarrow$iu & 14.3 & 14.5 & 14.9 & 14.5 & 14.6 & 15.1 \\
en$\rightarrow$ja & 14.4 & 14.5 & 14.5 & 15.0 & 14.5 & 15.6 \\
en$\rightarrow$ka & 11.0 & 12.5 & 11.5 & 11.5 & 12.1 & 12.9 \\
en$\rightarrow$kk & 4.5 & 4.7 & 4.9 & 3.8 & 5.0 & 5.2 \\
en$\rightarrow$km & 0.0 & 0.1 & 0.1 & 0.0 & 0.1 & 0.1 \\
en$\rightarrow$ko & 5.2 & 6.0 & 6.0 & 5.6 & 5.9 & 6.3 \\
en$\rightarrow$lt & 11.2 & 12.2 & 11.9 & 12.0 & 12.6 & 12.6 \\
en$\rightarrow$lv & 13.9 & 14.7 & 15.5 & 15.2 & 15.6 & 16.1 \\
en$\rightarrow$mk & 23.9 & 25.8 & 24.8 & 24.1 & 25.6 & 26.7 \\
en$\rightarrow$ml & 4.7 & 4.9 & 5.8 & 4.9 & 4.9 & 5.7 \\
en$\rightarrow$mn & 6.7 & 8.0 & 7.3 & 7.8 & 7.7 & 8.2 \\
en$\rightarrow$mr & 9.0 & 12.0 & 11.4 & 10.2 & 11.4 & 11.8 \\
en$\rightarrow$my & 21.5 & 22.4 & 21.9 & 21.9 & 22.5 & 23.0 \\
en$\rightarrow$ne & 6.1 & 5.9 & 5.4 & 6.2 & 6.4 & 5.8 \\
en$\rightarrow$nl & 26.5 & 29.4 & 29.8 & 28.5 & 29.6 & 30.6 \\
en$\rightarrow$pl & 19.8 & 20.6 & 20.6 & 20.8 & 21.4 & 21.4 \\
en$\rightarrow$ps & 6.1 & 6.5 & 7.0 & 6.6 & 6.6 & 7.6 \\
en$\rightarrow$pt & 34.4 & 37.8 & 37.6 & 36.7 & 38.2 & 39.0 \\
en$\rightarrow$ro & 22.1 & 23.7 & 24.2 & 23.5 & 24.2 & 24.5 \\
en$\rightarrow$ru & 22.4 & 23.1 & 23.7 & 24.0 & 24.4 & 24.1 \\
en$\rightarrow$si & 0.8 & 1.7 & 2.0 & 1.0 & 1.5 & 2.4 \\
en$\rightarrow$sl & 19.2 & 20.8 & 20.5 & 20.5 & 21.4 & 21.8 \\
en$\rightarrow$sv & 30.5 & 34.9 & 35.6 & 33.3 & 34.8 & 35.9 \\
en$\rightarrow$ta & 6.3 & 6.6 & 6.8 & 6.4 & 6.9 & 7.0 \\
en$\rightarrow$te & 21.5 & 24.1 & 25.9 & 21.9 & 22.8 & 23.9 \\
en$\rightarrow$th & 16.8 & 18.6 & 19.2 & 17.4 & 18.3 & 19.7 \\
en$\rightarrow$tr & 14.1 & 15.5 & 16.5 & 15.5 & 16.2 & 16.8 \\
en$\rightarrow$uk & 20.3 & 21.6 & 21.2 & 21.5 & 21.9 & 22.3 \\
en$\rightarrow$ur & 13.9 & 17.9 & 18.7 & 14.6 & 16.5 & 18.9 \\
en$\rightarrow$vi & 27.1 & 28.6 & 29.2 & 28.3 & 28.6 & 30.0 \\
en$\rightarrow$xh & 12.4 & 12.7 & 12.2 & 12.8 & 12.8 & 12.9 \\
en$\rightarrow$zh & 24.1 & 25.6 & 25.9 & 25.7 & 26.4 & 26.5 \\
\bottomrule[1.5pt]

\end{tabular}
    \caption{BLUE ($\uparrow$) scores of the \textit{Transformer-base} models on the \textbf{en$\rightarrow$X} pairs of ML50.}
	\label{table:random-ml50-base-en-many}
\end{table*}
\endgroup

\begingroup
\setlength{\tabcolsep}{8.5pt} %
\renewcommand{\arraystretch}{1.2} %
\begin{table*}[!tb]
	\small
	\centering
\begin{tabular}{lrrrrrr}
\toprule[1.5pt]

\multicolumn{1}{c}{\multirow{2}[2]{*}{\textbf{Language}}} & 
\multicolumn{1}{c}{\multirow{2}[2]{*}{\textbf{Transformer-Base}}} & 
\multicolumn{5}{c}{\textbf{+adapters}} \\ 
\cmidrule(l){3-7} 
\multicolumn{1}{c}{} & 
\multicolumn{1}{c}{} & 
\textbf{lang} & 
\textbf{pair} & 
\textbf{hyper (base)} & 
\textbf{hyper (base)} & 
\textbf{hyper (base)} \\ \midrule

a$\rightarrow$en & 27.4 & 26.0 & 26.0 & 30.3 & 29.1 & 27.5 \\
ar$\rightarrow$en & 29.8 & 32.7 & 32.6 & 31.1 & 31.9 & 33.0 \\
az$\rightarrow$en & 15.1 & 15.1 & 14.7 & 15.8 & 16.1 & 15.3 \\
bn$\rightarrow$en & 17.6 & 17.9 & 16.5 & 19.6 & 18.8 & 19.9 \\
cs$\rightarrow$en & 26.5 & 27.3 & 26.4 & 27.5 & 27.5 & 27.5 \\
de$\rightarrow$en & 35.6 & 37.1 & 36.6 & 36.4 & 36.7 & 37.4 \\
es$\rightarrow$en & 29.1 & 30.5 & 27.5 & 29.1 & 29.6 & 29.0 \\
et$\rightarrow$en & 22.8 & 24.2 & 23.1 & 24.0 & 24.3 & 24.8 \\
fa$\rightarrow$en & 27.3 & 30.5 & 29.3 & 28.2 & 28.6 & 30.6 \\
fi$\rightarrow$en & 22.7 & 24.9 & 23.9 & 24.2 & 24.8 & 25.1 \\
fr$\rightarrow$en & 33.9 & 34.6 & 34.0 & 34.8 & 34.9 & 35.0 \\
gl$\rightarrow$en & 34.7 & 33.7 & 33.5 & 35.9 & 35.7 & 34.5 \\
gu$\rightarrow$en & 2.3 & 1.1 & 0.9 & 2.4 & 2.3 & 2.2 \\
he$\rightarrow$en & 35.4 & 38.6 & 37.9 & 36.1 & 36.8 & 38.4 \\
hi$\rightarrow$en & 20.5 & 20.9 & 20.4 & 21.4 & 21.2 & 21.8 \\
hr$\rightarrow$en & 37.0 & 40.1 & 39.6 & 38.5 & 38.9 & 39.9 \\
id$\rightarrow$en & 31.5 & 34.9 & 34.5 & 33.1 & 33.7 & 34.7 \\
it$\rightarrow$en & 36.9 & 39.5 & 39.2 & 38.2 & 39.0 & 39.9 \\
iu$\rightarrow$en & 24.5 & 26.7 & 26.4 & 25.1 & 25.7 & 27.3 \\
ja$\rightarrow$en & 14.4 & 15.6 & 15.2 & 15.6 & 15.6 & 15.8 \\
ka$\rightarrow$en & 23.2 & 23.3 & 21.5 & 23.0 & 23.6 & 23.6 \\
kk$\rightarrow$en & 13.3 & 12.8 & 12.1 & 12.5 & 13.4 & 13.3 \\
km$\rightarrow$en & 6.3 & 6.7 & 5.7 & 6.3 & 6.0 & 7.3 \\
ko$\rightarrow$en & 15.6 & 17.2 & 16.9 & 16.5 & 16.3 & 17.4 \\
lt$\rightarrow$en & 25.7 & 28.0 & 27.6 & 26.9 & 27.1 & 27.8 \\
lv$\rightarrow$en & 17.9 & 19.1 & 19.0 & 19.0 & 19.2 & 19.2 \\
mk$\rightarrow$en & 35.6 & 37.0 & 36.1 & 36.9 & 36.8 & 37.0 \\
ml$\rightarrow$en & 13.5 & 16.0 & 15.7 & 14.0 & 14.5 & 15.5 \\
mn$\rightarrow$en & 10.5 & 11.0 & 11.3 & 10.8 & 11.9 & 11.6 \\
mr$\rightarrow$en & 12.8 & 13.1 & 12.3 & 13.7 & 13.7 & 14.0 \\
my$\rightarrow$en & 24.6 & 25.5 & 24.2 & 25.7 & 25.6 & 26.2 \\
ne$\rightarrow$en & 15.7 & 13.9 & 13.6 & 15.8 & 15.3 & 14.2 \\
nl$\rightarrow$en & 32.9 & 35.1 & 34.8 & 33.9 & 34.3 & 35.4 \\
pl$\rightarrow$en & 25.9 & 26.4 & 26.1 & 26.7 & 27.2 & 26.7 \\
ps$\rightarrow$en & 11.0 & 10.8 & 13.2 & 12.2 & 12.4 & 12.8 \\
pt$\rightarrow$en & 42.4 & 44.8 & 44.1 & 44.0 & 44.3 & 45.2 \\
ro$\rightarrow$en & 31.5 & 34.1 & 33.5 & 32.9 & 33.7 & 34.8 \\
ru$\rightarrow$en & 34.2 & 35.2 & 34.8 & 34.5 & 35.1 & 35.3 \\
si$\rightarrow$en & 8.4 & 10.4 & 10.0 & 8.8 & 9.3 & 10.3 \\
sl$\rightarrow$en & 28.1 & 28.4 & 28.7 & 29.6 & 28.9 & 29.8 \\
sv$\rightarrow$en & 39.9 & 43.2 & 42.6 & 41.2 & 42.3 & 43.1 \\
ta$\rightarrow$en & 15.1 & 16.0 & 15.6 & 16.2 & 16.2 & 15.9 \\
te$\rightarrow$en & 30.3 & 33.8 & 32.8 & 30.9 & 31.7 & 33.8 \\
th$\rightarrow$en & 23.8 & 25.8 & 24.8 & 24.2 & 24.7 & 25.5 \\
tr$\rightarrow$en & 18.4 & 20.4 & 20.2 & 19.5 & 19.6 & 20.3 \\
uk$\rightarrow$en & 30.0 & 31.6 & 31.2 & 30.7 & 31.3 & 32.0 \\
ur$\rightarrow$en & 22.4 & 24.6 & 24.5 & 22.3 & 23.8 & 25.0 \\
vi$\rightarrow$en & 26.4 & 28.1 & 27.8 & 27.5 & 27.8 & 28.0 \\
xh$\rightarrow$en & 12.2 & 11.7 & 11.5 & 12.0 & 12.6 & 12.6 \\

\bottomrule[1.5pt]

\end{tabular}
    \caption{BLUE ($\uparrow$) scores of the \textit{Transformer-base} models on the \textbf{X$\rightarrow$en} pairs of ML50.}
	\label{table:random-ml50-base-many-en}
\end{table*}
\endgroup

\end{document}